\documentclass[journal,twoside]{IEEEtran}
\usepackage{amsmath,amsfonts}
\usepackage{algorithmic}
\usepackage{algorithm}
\usepackage{array}
\usepackage{textcomp}
\usepackage{url}
\usepackage{verbatim}
\usepackage{graphicx}
\usepackage[caption=false,font=footnotesize]{subfig}
\usepackage{cite}
\usepackage{booktabs}
\usepackage{adjustbox}
\usepackage{multirow}
\usepackage[section]{placeins}
\usepackage[percent]{overpic}
\usepackage{balance}
\usepackage{xcolor}
\usepackage{siunitx}
\usepackage{fancyhdr}
\usepackage{hyperref}

\begin{document}

\title{
Static and Dynamic Representations for Tactile Contact-Angle Estimation with Event-Based Sensors
}

\author{Yanhui Lu, 
        Efi Psomopoulou,~\IEEEmembership{Member,~IEEE}, 
        Benjamin~Ward-Cherrier,~\IEEEmembership{Member,~IEEE}%

\thanks{This work was partially supported by the Horizon Europe research and innovation program (MANiBOT, Grant 101120823), the Royal Society International Collaboration Awards (South Korea), ARIA on Robot Dexterity and the Royal Academy of Engineering Fellowship (Grant RF02021071).}%
\thanks{Y. Lu, E. Psomopoulou, and B. Ward-Cherrier are with the School of Engineering Mathematics and Technology, University of Bristol, U.K.}%
\thanks{Corresponding author: Yanhui Lu (email: mb23300@bristol.ac.uk).}%
}

\maketitle
\thispagestyle{fancy}

\bstctlcite{myIEEEctl}

\begin{abstract}

Event-based tactile sensing offers low-latency signal acquisition for contact-rich robotic interaction. This paper investigates contact-angle estimation using event streams from an event-based tactile sensor (NeuroTac) and compares three event-derived spatial contour representations: a dynamic representation capturing recent event activity, a static representation recovering a more persistent contact state, and their combined representation. Across the evaluated motion scenarios, all representation pipelines exhibited P99 processing latency below \boldmath$10\,\text{ms}$\unboldmath\ at all tested sampling intervals, demonstrating their potential for high-frequency event-based tactile angle estimation in robotic manipulation. The static representation consistently achieved marginally better performance than the dynamic and combined representations under scenario-specific training, yielding a mean overall MAE of \boldmath$0.160^\circ$\unboldmath\ during continuous sensor rolling and a stop-phase mean MAE of \boldmath$0.251^\circ$\unboldmath\ during randomly inserted motion interruptions. It also exhibited smaller performance fluctuations across speed and indentation depth variations than the other two representations.

\end{abstract}



\section{Introduction}

\IEEEPARstart{I}{n} the human tactile system, there are two major types of mechanoreceptors, namely rapidly adapting (RA) and slowly adapting (SA) receptors~\cite{abraira2013sensory}. RA receptors primarily encode instantaneous changes during contact, whereas SA receptors emphasize the stable spatial distribution and deformation cues under sustained contact. The complementary roles of these two pathways jointly support the broad adaptability of human touch across different scenarios, including shape perception, slip response, and manipulation controls~\cite{o2021review}.

For vision-based tactile sensors (VBTSs), contact deformation states are typically reflected by the spatial distribution of internal pins or markers on the camera imaging plane. With the development of neuromorphic technology, a class of event-based tactile sensors based on vision sensing, often referred to as neuromorphic vision-based tactile sensors (NVBTSs)~\cite{ward2020neurotac,funk2024evetac} have attracted increasing attention. These sensors introduce the asynchronous event generation and output mechanism of event cameras into tactile sensing, such that local contact changes on the sensor surface are encoded as sparse and asynchronous events to represent the contact process~\cite{chakravarthi2024recent}. Compared with conventional frame-based tactile imaging, this change-driven signal-generation mechanism offers advantages including low power consumption, high temporal resolution, and high dynamic range, while avoiding the exposure latency inherent in frame-based cameras~\cite{huang2022real}. It has shown efficient and accurate sensing performance in a variety of tactile tasks centered on contact changes, such as incipient slip detection~\cite{lu2026neuromorphic}, and tactile perception related to micro-vibrations or textures~\cite{funk2024evetac,brayshaw2024neuromorphic}. However, in human tactile perception, stable information encoded by slow-adapting receptors, such as pressure distribution and skin deformation, is also important because it provides more consistent cues for contact-state perception~\cite{woo2015merkel,birznieks2009slowly,watkins2022slowly}. Therefore, both dynamic and static cues deserve attention in event-based tactile perception.

\begin{figure}[!t]
    \centering
    \includegraphics[width=\columnwidth,trim=5pt 5pt 5pt 5pt,clip]{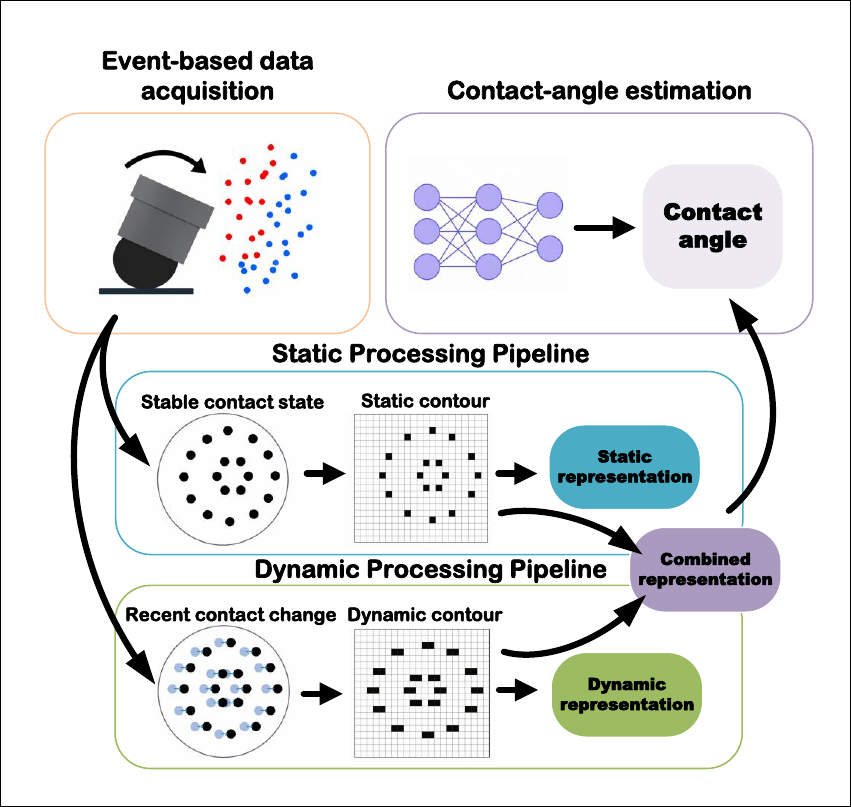}
    \caption{System-level overview of the proposed framework, where the same tactile event stream is processed through static and dynamic pipelines inspired by human SA and RA receptors.}
    \label{fig:top}
\end{figure}

Contact angle is one of the key variables for describing tactile contact states. In contact-rich manipulation, it can reflect the underlying contact mechanics and their variation trends~\cite{van2015learning,hogan2020tactile}, thereby supporting control decisions related to contact alignment, grasp stability, and posture adjustment~\cite{ota2023tactile}. However, during continuous sensor pose variation, how to represent event streams for contact-angle estimation remains underexplored, particularly with respect to whether change-driven or state-driven information is more critical for this task. Motivated by the distinction between rapidly adapting and slowly adapting receptor signals in the human tactile system, this work constructs a static and a dynamic representation from the event-based data stream and compares their performance in contact-angle regression with a fully-connected 3-layer neural network. The main contributions of this work are as follows:

\begin{itemize}

    \item Inspired by the functional distinction between rapidly adapting (RA) and slowly adapting (SA) mechanoreceptors, we design two separate pipelines to process event-based tactile data: a dynamic representation emphasizing short-term changes, and a static representation emphasizing sustained contact states.

    \item We demonstrate high-accuracy (MAE = $0.160^{\circ}$), low-latency ($<10$\,ms) contact-angle estimation with event-based tactile sensing during unidirectional continuous sensor rolling under varying speed, indentation depth, and motion-interruption conditions.

    \item We compare static, dynamic, and their combined representation pipelines, showing that the static representation provides the most accurate and stable contact-angle estimation performance.
    
\end{itemize}

\section{Related work}

Event-based tactile sensing has been explored in a growing range of robotic perception tasks, including slip detection~\cite{lu2026neuromorphic}, material recognition~\cite{brayshaw2024neuromorphic}, and fault detection~\cite{sherif2025real}. In these applications, raw event streams are typically segmented into fixed-time or event-count windows, and then aggregate the events within each window through operations such as event accumulation or voxelization to form tensor-like inputs for network processing. In this work, we adopt this window-based event aggregation strategy and use it to integrate short-term event activity into a spatial contour distribution that reflects recent contact variations, which is treated as the dynamic representation.

To obtain a more stable representation from event streams and better recover persistent scene structure, prior studies have explored learning-based reconstruction models that convert sparse, asynchronous event signals into image-like outputs similar to those of conventional frame cameras. For example, \cite{rebecq2019events} used a recurrent convolutional network to reconstruct intensity images directly from event streams; \cite{paredes2021back} learned event-to-image reconstruction in a self-supervised manner under constraints such as photometric constancy; and \cite{yang2023learning} further fused event signals with APS frames to alleviate exposure inconsistencies and ghosting artifacts. Although these methods can recover spatial structures effectively under favourable conditions and provide a certain degree of noise correction, they usually require additional training and incur relatively high inference costs. For NVBTSs applications that emphasize low-latency and low-power processing, such characteristics may limit their practical efficiency. Moreover, methods that rely on sufficiently dense events or additional frame data to ensure stable and clear reconstruction may become unreliable when the event rate is low.

Unlike learning-based methods, non-learning-based methods may require task-specific denoising to reduce structured background noise, but they are generally simpler to implement and better suited to rapid recovery of stable state-like spatial information. These methods can be broadly divided into two categories. One class represents local spatiotemporal structure by maintaining the timestamp of the most recent event at each pixel together with its decay history~\cite{mueggler2015lifetime,lagorce2016hots}. The other continuously integrates event streams into temporally updated spatial representations through decaying continuous-time filtering mechanisms~\cite{scheerlinck2018continuous}. In this paper, we adopt the continuous-time decaying integration framework in~\cite{scheerlinck2018continuous} to fuse event streams into a stable representation, which serves as the static pipeline for contact-angle regression.

For contact-angle estimation in tactile sensing, early studies mainly relied on image data from conventional frame-based VBTSs, exploiting static skin-deformation cues and using regression networks to extract geometric features for angle estimation~\cite{zaid2022elastomer,halwani2024novel}. In the event-based tactile domain, earlier studies have also shown that event streams from NVBTSs can support contact-edge angle prediction when the task is formulated as a discrete classification problem~\cite{macdonald2022neuromorphic}. More closely related to our work, \cite{sajwani2023tactigraph} adopted a more event-native processing framework by feeding the event stream within a temporal window into a graph convolutional network, and performing roll/pitch angle estimation. However, its performance remains sensitive to the number of events contained in the input window, and its modelling strategy mainly emphasizes short-term dynamic event activities. In addition, the relatively large network size limits its suitability for low-latency processing. Motivated by these limitations, this study focuses on a lightweight angle regression framework for contact-angle estimation during continuous rolling. Furthermore, we systematically compare static, dynamic, and combined representations to analyse their respective performance in this task.

\section{Methods}

\begin{figure*}[!t]
    \centering
    \includegraphics[width=\textwidth,trim=5pt 5pt 5pt 5pt,clip]{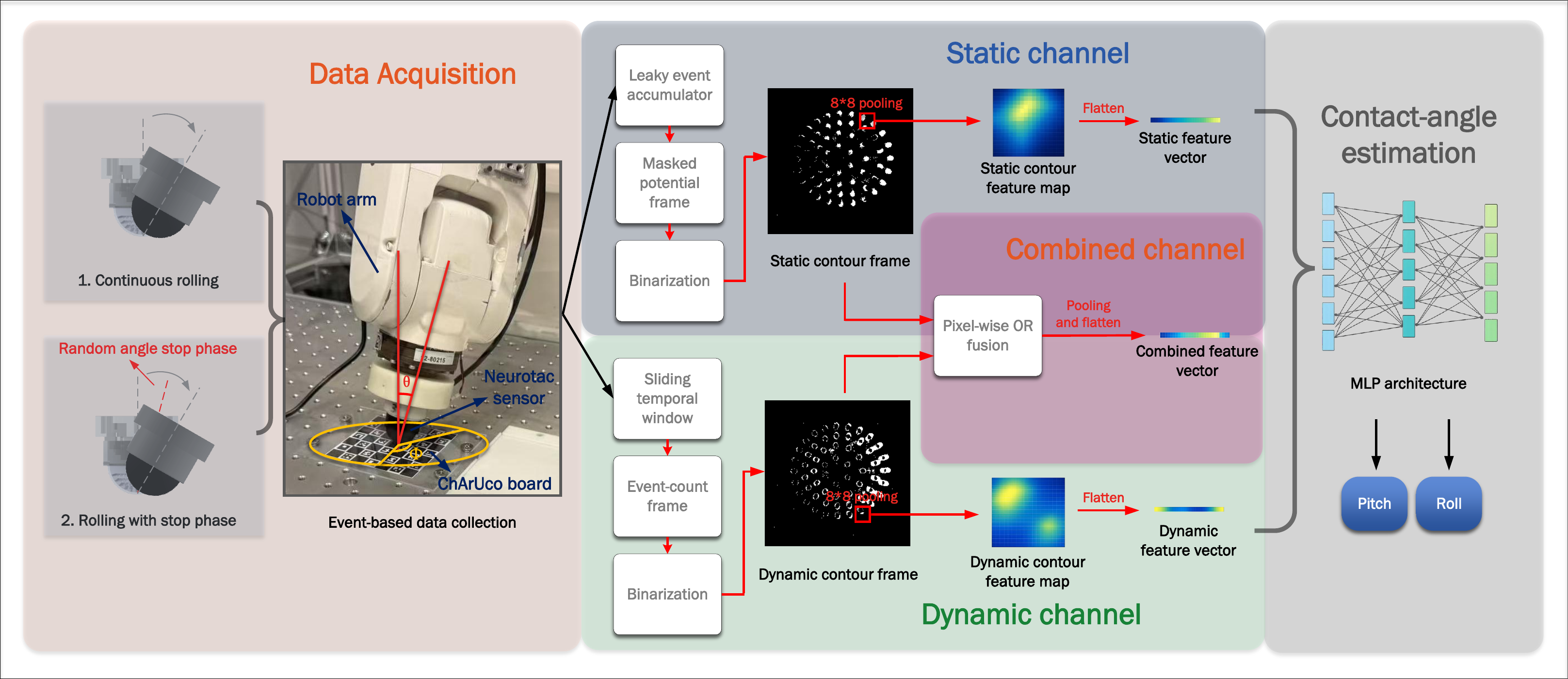}
    \caption{Implementation-level framework of the proposed system. Event-based tactile data are collected using a NeuroTac sensor mounted on a robot arm under continuous rolling and rolling with a randomly inserted stop phase. The event stream is processed by static and dynamic pipelines to generate corresponding spatial contour representations, and pixel-wise OR fusion is used to form a combined representation. The three representations are separately fed into an MLP for contact-angle estimation.}
    \label{fig:overall_top}
\end{figure*}

\subsection{Data acquisition and label generation}

To acquire pose-angle data during continuous rolling, we used the data-collection setup shown in Fig.~\ref{fig:overall_top}. A NeuroTac sensor \cite{ward2020neurotac} was mounted on the end-effector of a 6-dof robotic arm (ABB, IRB120) to capture event-based tactile data. The event camera within the NeuroTac (Synsense, DVXplorer) includes an IMU whose timestamps are synchronized with the output event timestamps, enabling pose recording for label generation. The raw event stream was cropped to a centered \(240 \times 240\) region for further processing.

The robot was controlled in spherical coordinates, where $\theta$ and $\phi$ denote the tilt magnitude and tilt direction, respectively. In each trial, $\phi$ was selected from 20 uniformly distributed azimuthal directions around the circle, while $\theta$ increased continuously from $0^\circ$ to $20^\circ$ at a constant angular speed and indentation depth.

The data collection was organized into three experiments. Experiment 1 was designed to evaluate the performance of different representations under continuous sensor rolling motion. For each of the 20 directions, data were collected under five rolling speeds from $2$ to $10\,^\circ/\text{s}$ with a step of $2\,^\circ/\text{s}$, and five evenly spaced indentation depths from $1.0$ to $3.0\,\text{mm}$. Two trials were collected for each speed--depth--direction combination. Experiment 2 was designed to evaluate performance under continuous motion with randomly inserted motion interruptions. In Experiment 2, a $500\,\text{ms}$ stop phase was inserted at a random $\theta$ angle during each trial, while the other settings remained the same as in Experiment 1. The trials from Experiments 1 and 2 were treated as two separate datasets, each of which was split into training, validation, and test sets using the same $70{:}15{:}15$ ratio, with the validation and test trials randomly selected to maintain approximately balanced marginal distributions of speed, depth, and direction.

Experiment 3 was designed to evaluate model generalization across both training and interpolated unseen speeds and indentation depths. For speed-wise evaluation, 30 trials were collected at each target speed using a fixed set of depth--direction combinations across all speeds. For depth-wise evaluation, 30 trials were collected at each target depth using a fixed set of speed--direction combinations across all depths. In both cases, the combinations were chosen from values present in the training set to maintain approximately balanced marginal distributions over the remaining variables. This design enables a fairer assessment of the individual effect of speed or depth.

For label generation, we followed the calibration framework in \cite{halwani2024novel,sajwani2023tactigraph}. First, static frames reconstructed by the E2VID model \cite{rebecq2019high} were used together with a ChArUco board for camera intrinsic calibration. The required extrinsic transformations were then obtained through geometric calibration. During the formal data-collection trials, the event-based tactile sensor's IMU measurements were used to estimate the tilt magnitude $\theta$. The axial angular velocity was refined using the commanded speed and motion conditions as priors, then bias-compensated, scale-calibrated, zero-phase filtered, and integrated to recover the sensor tilt-magnitude trajectory $\theta(t)$. The commanded angles at randomly inserted intermediate stops were used for validation, yielding an error of approximately $-0.01^\circ \pm 0.04^\circ$ between the integrated tilt magnitude and the robot-commanded stop angle. The recovered $\theta(t)$ and commanded tilt direction $\phi$ were combined with the calibrated extrinsics and sensor geometry to obtain the sensor-to-plane contact-angle labels.

\subsection{Representation processing}

To enable a fair comparison between the static, dynamic and combined representation pipelines, we applied the same regression backend and training protocol to all three. For each representation, we first obtained binary frames to reduce the effects of pixel-intensity variations and accumulated noise events. We then applied an $8 \times 8$ mean-pooling operation to the central $240 \times 240$ region, resulting in a $30 \times 30$ feature map which was flattened and fed into the network. The detailed procedures for generating the binary frames of each representation are as follows.

\subsubsection{Dynamic representation}
For the representation that preserves dynamic information, we adopted a conventional time-window slicing strategy on the event stream. For each sampled data point in a trial with timestamp $t_k$, we extracted a fixed-length event window preceding it, i.e.,
\begin{equation}
[t_k - \Delta t,\; t_k],
\end{equation}
where $\Delta t$ denotes the window length (in ms). All positive and negative events within the window were accumulated into an event-count frame which was then binarized using an event-count threshold. To improve computational efficiency, we adopted an incremental sliding-window update strategy, in which the count frame was initialized once and then updated by adding newly entered events and removing expired events as the window advances.

\subsubsection{Static representation}
For the static representation, events were incrementally fed into an accumulator from the start of each trial up to each sampled timestamp $t_k$, producing a time-evolving pixel potential state. For any pixel, its potential at time $t_k$ is mathematically expressed as:

\begin{equation}
V_u(t_k)
= V_u(\tilde{t}_u)\exp\!\left(-\frac{t_k-\tilde{t}_u}{\tau}\right) + \eta\,p_k .
\end{equation}

where $V_u(t)$ denotes the potential (state) at pixel $u$ at time $t$; $\eta$ is the event contribution per event; $\tau$ is the exponential decay time constant; an event occurring at time $t_k$ is denoted by $e_k=(u,t_k,p_k)$, where the polarity satisfies $p_k\in\{+1,-1\}$; $\tilde{t}_u$ is the timestamp when pixel $u$ was last updated (i.e., the time of its most recent event).

The decay term of a pixel was computed and updated only when that pixel received a new event, rather than being refreshed at every time step. This allowed the accumulator to preserve the previous state representation when the event rate was extremely low while reducing computational cost. 

For each data point at time $t_k$, we read the pixel potential field accumulated up to $t_k$ and reconstructed it as a static frame. We then used a frame gathered by the sensor during the no-contact condition as a mask, set the pixels at its non-zero locations in the static frame to zero, and binarized the result using an event-contribution threshold. The resulting frame represents the sensor state induced by the contact condition at time $t_k$.

\subsubsection{Combined representation}
For the combined representation, we fused the binary frames generated by the static and dynamic pipelines at each sampled timestamp $t_k$. Specifically, after the static and dynamic frames were separately binarized, a pixel-wise OR operation was applied to combine them into a single binary frame. In this way, a pixel in the combined frame was set to 1 if it was activated in either the static or dynamic representation, and was set to 0 only when it was inactive in both. The resulting combined binary frame has the same spatial size as each individual representation.

\subsection{Network training and evaluation}

For Experiments 1 and 3, all representation pipelines used the same data extraction setup: each trial was first sampled at a $1\,\text{ms}$ interval from motion onset to motion end, and the same 50 equally spaced data points were extracted from the motion phase. Experiment 2 used the same procedure, except that 10 additional equally spaced data points were extracted from the randomly inserted stop phase of each trial.

Processing latency was evaluated using the 150 continuous-motion test trials from Experiment 1. For each sampling interval, every trial was converted into a sampled sequence and processed sequentially in temporal order. After processing, 50 uniformly spaced points were selected from each trial, and their corresponding processing latencies were used to characterize the latency distribution under each sampling interval, thereby avoiding over-representation of longer trials.

The experimental protocol was as follows. In Experiment 1, we performed a grid search over all combinations of the key hyperparameters for the static and dynamic representations (decay coefficient and binarization threshold for the static representation; temporal window length and binarization threshold for the dynamic representation). For each hyperparameter configuration, the checkpoint with the best validation performance was first selected. These best checkpoints were then compared across all configurations, and the hyperparameter setting with the best overall validation performance was chosen. For computational efficiency, this hyperparameter search was conducted using a single training run for each configuration. After the optimal hyperparameters were selected, the final quantitative comparison among the static, dynamic, and combined representations was repeated over ten independent training runs, and the results were reported as the mean and standard deviation. In Experiment 2, we used the optimal hyperparameters obtained in Experiment 1 and retrained the model on the new dataset. In Experiment 3, for each representation, we directly used the ten independently trained models obtained from Experiment 1 to evaluate performance on datasets collected under different speeds and depths.

A three-layer fully connected Multi-layer perceptron was used for fast angle regression. The sizes of the three hidden layers were 256, 128, and 64, respectively. We trained each model with the SmoothL1 (Huber) loss ($\beta=1.0$) and the AdamW optimizer ($\mathrm{lr}=5\times10^{-3}$, weight decay $10^{-4}$), using a batch size of $32$ for $100$ epochs.


\section{Results}

\subsection{Angle regression during continuous motion}
\label{sec:exp1}

\begin{figure}[t]
    \centering

    \subfloat[]{%
        \includegraphics[width=0.48\linewidth]{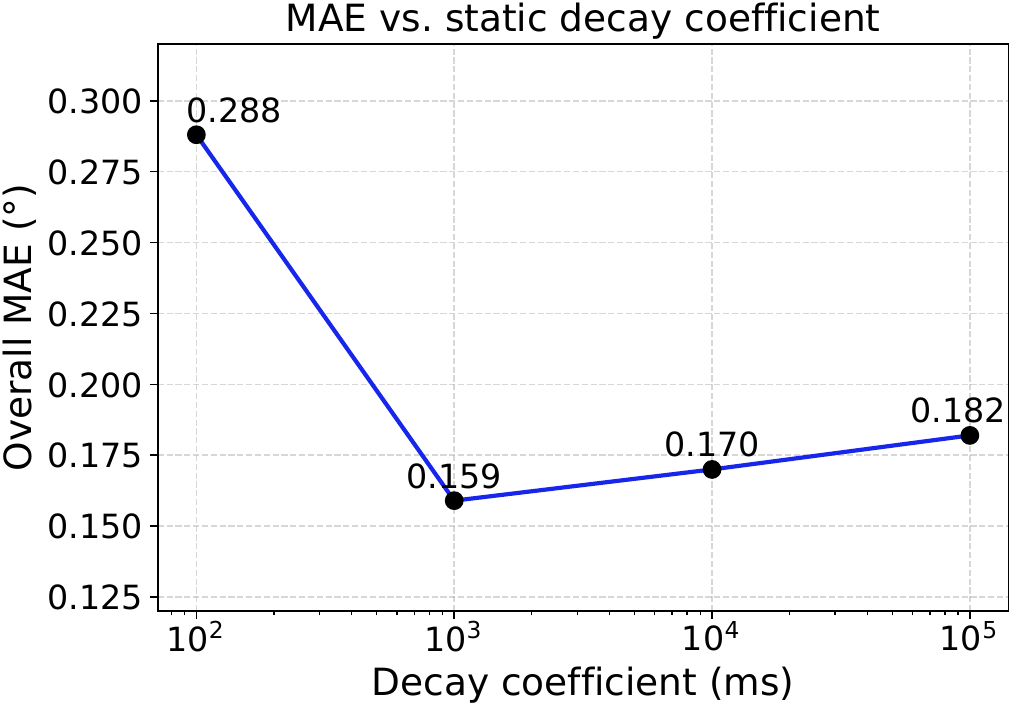}
        \label{fig:static_decay}
    }
    \hfill
    \subfloat[]{%
        \includegraphics[width=0.48\linewidth]{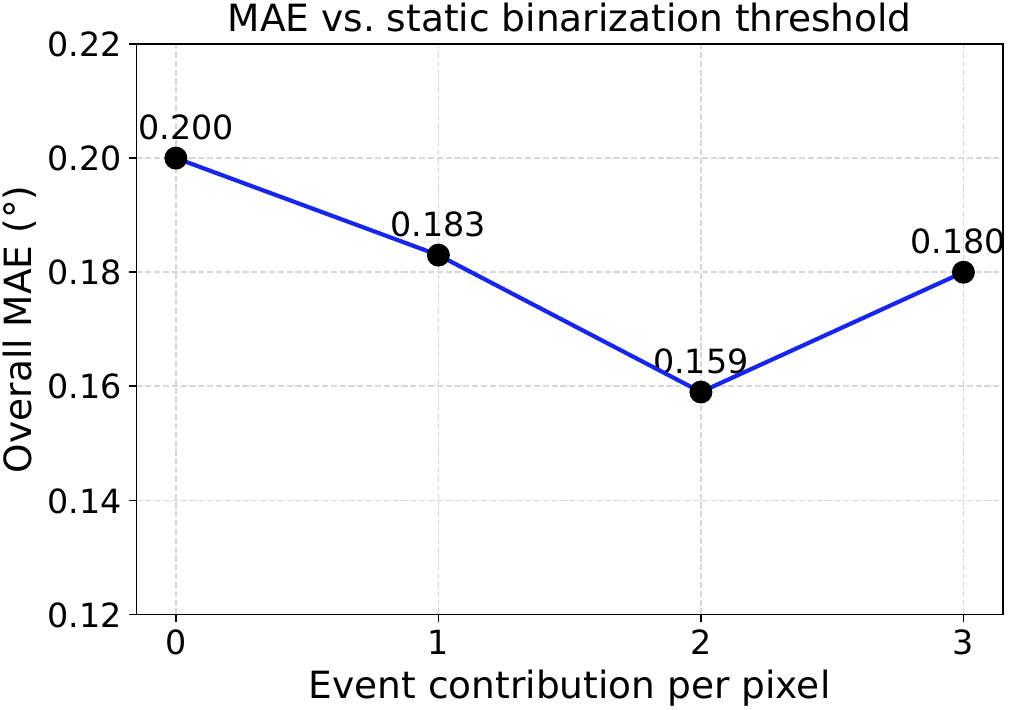}
        \label{fig:static_threshold}
    }

    \vspace{0.5em}

    \subfloat[]{%
        \includegraphics[width=0.48\linewidth]{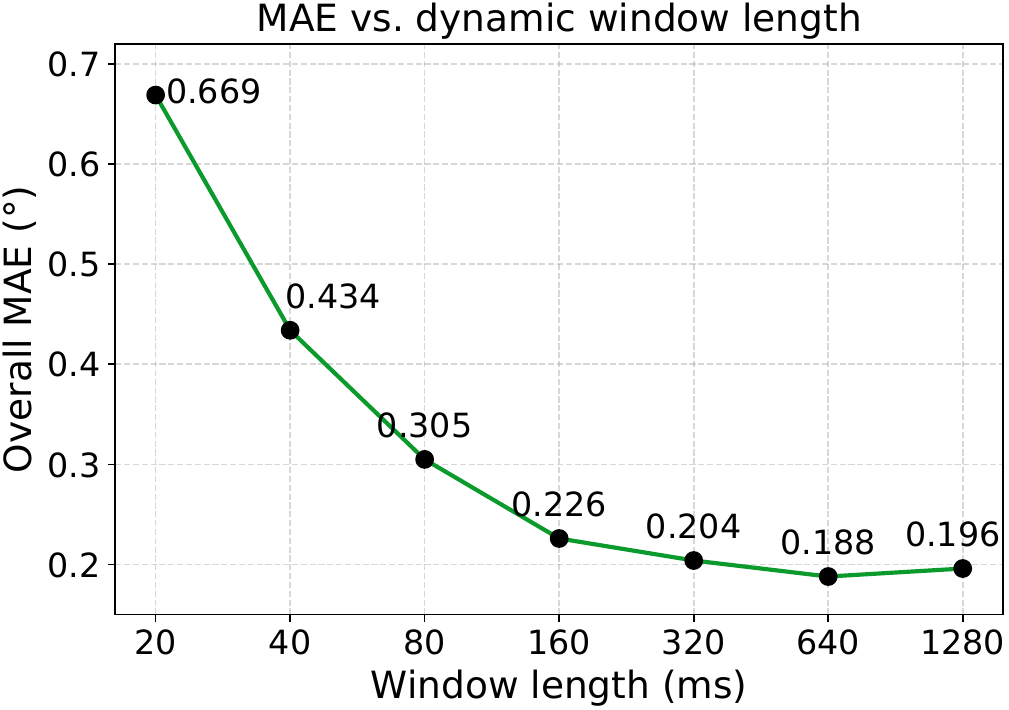}
        \label{fig:dynamic_window}
    }
    \hfill
    \subfloat[]{%
        \includegraphics[width=0.48\linewidth]{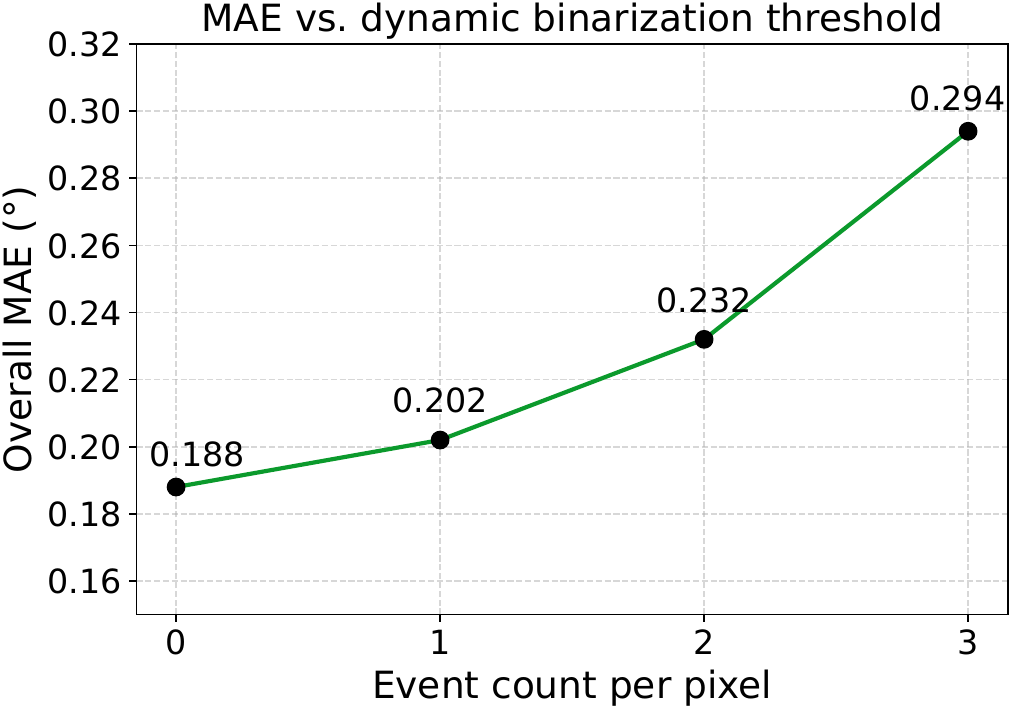}
        \label{fig:dynamic_threshold}
    }

    \caption{Parameter tuning results for the static and dynamic representations on the validation set.}
    \label{fig:parameter_tuning}
\end{figure}

Fig.~\ref{fig:parameter_tuning} presents the tuning results by varying one hyperparameter while fixing the other. For the static representation, with the binarization threshold fixed at 2 event contributions per pixel, the lowest validation regression error was achieved with a decay coefficient of $1000\,\text{ms}$, whereas the shorter coefficient of $100\,\text{ms}$ produced substantially larger errors. When the decay coefficient was fixed at $1000\,\text{ms}$, a binarization threshold of 0 preserved all noise responses in the contour and degraded regression accuracy. As the threshold increased, noisy pixels were progressively suppressed and the performance improved, with the best result obtained at 2 contributions per pixel.

For the dynamic representation, with the binarization threshold fixed at 0 event count, increasing the temporal window length initially reduced the validation regression error rapidly, after which the performance gradually stabilized from around $320\,\text{ms}$, with the lowest error observed at $640\,\text{ms}$ but only by a small margin. When the window length was fixed at $640\,\text{ms}$, the best result was achieved with a threshold of 0, indicating that under continuous motion the dynamic representation was less affected by long-term noise accumulation than the static representation. Increasing the threshold further removed valid contour information together with noise and reduced regression accuracy.

\begin{table}[t]
\centering
\caption{Comparison of the best regression results under different representation settings. Values are reported as mean and standard deviation.}
\label{tab:representation_comparison}
\begin{tabular}{lccc}
\hline
Representation & Overall MAE ($^\circ$) & Pitch MAE ($^\circ$) & Roll MAE ($^\circ$) \\
\hline
Static   & $0.160 \pm 0.004$ & $0.158 \pm 0.005$ & $0.163 \pm 0.008$ \\
Dynamic  & $0.186 \pm 0.007$ & $0.183 \pm 0.007$ & $0.189 \pm 0.010$ \\
Combined & $0.173 \pm 0.006$ & $0.169 \pm 0.008$ & $0.177 \pm 0.006$ \\
\hline
\end{tabular}
\end{table}

Table~\ref{tab:representation_comparison} reports the test-set regression errors of the three representations under their respective optimal hyperparameter settings, with results presented as the mean and standard deviation over ten training runs. The comparison indicates that the performance differences among the three representation settings are relatively small. During continuous motion, the static representation yielded slightly lower mean regression errors than the dynamic representation. Moreover, combining the two representations did not further improve the estimation results.

\subsection{Angle regression under interruption of motion}

\begin{figure}[!ht]
    \centering

    \subfloat[Motion phase MAE comparison.]{%
        \includegraphics[width=0.48\linewidth]{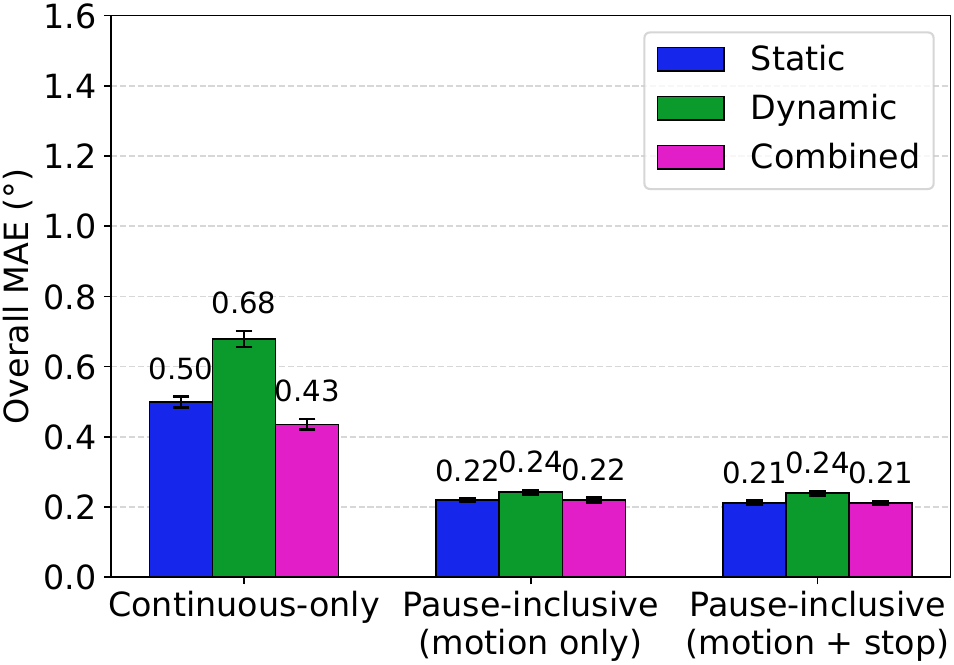}
        \label{fig:motion_phase_mae}
    }
    \hfill
    \subfloat[Stop phase MAE comparison.]{%
        \includegraphics[width=0.48\linewidth]{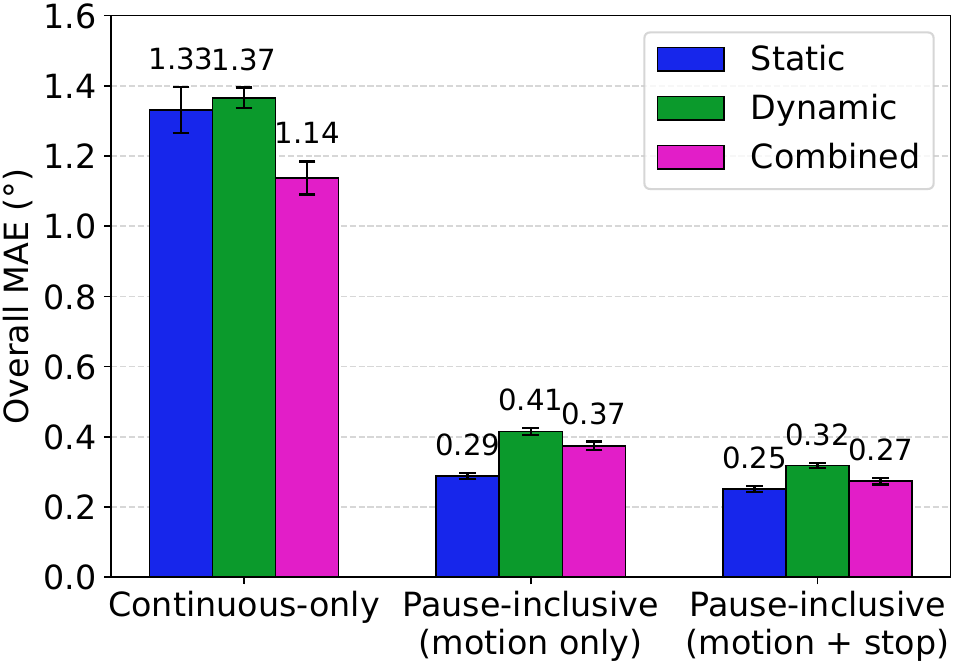}
        \label{fig:stop_phase_mae}
    }

    \caption{Comparison of regression errors under different representations and training settings, reported as mean and standard deviation over ten independent training runs. The three training settings are as follows: Continuous-only, trained using data extracted from pure continuous-motion trials; Pause-inclusive (motion only), trained using only the motion-phase data from pause-inclusive trials; and Pause-inclusive (motion + stop), trained using both motion- and stop-phase data from pause-inclusive trials.}
    \label{fig:motion_stop_mae_comparison}
\end{figure}

After introducing a short random angular pause into the continuous rolling process, the results in Fig.~\ref{fig:motion_stop_mae_comparison} show that the model in Section~\ref{sec:exp1} trained only on pure continuous-motion trials produced substantially larger mean regression errors in both the motion and stop phases than the other two training settings. This suggests that the pause may have introduced an unseen data pattern that was absent under purely continuous motion, likely due to internal dynamics of the silicone inside the sensor. In contrast, training with pause-inclusive trials markedly improved the estimation accuracy.

\begin{figure}[!t]
    \centering

    \subfloat[]{%
        \includegraphics[width=0.48\columnwidth]{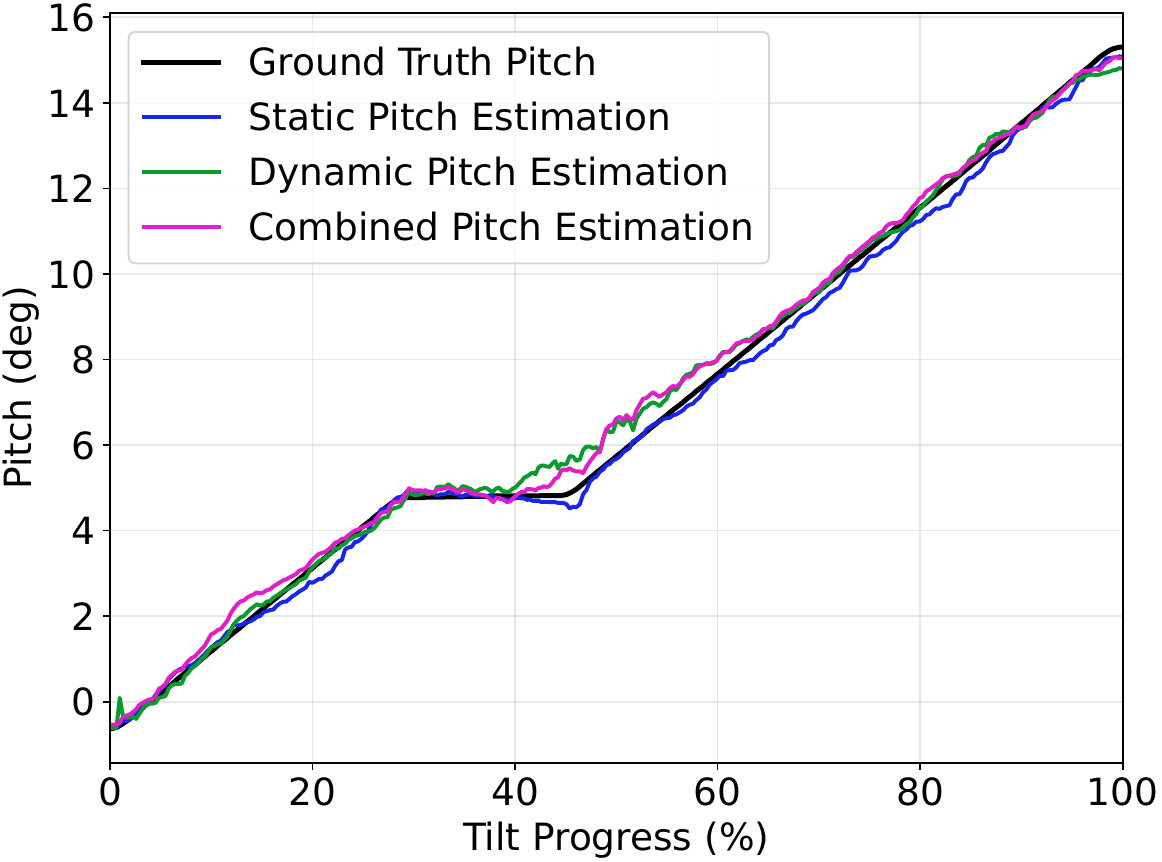}
        \label{fig:pause_pitch_example}
    }
    \hfill
    \subfloat[]{%
        \includegraphics[width=0.48\columnwidth]{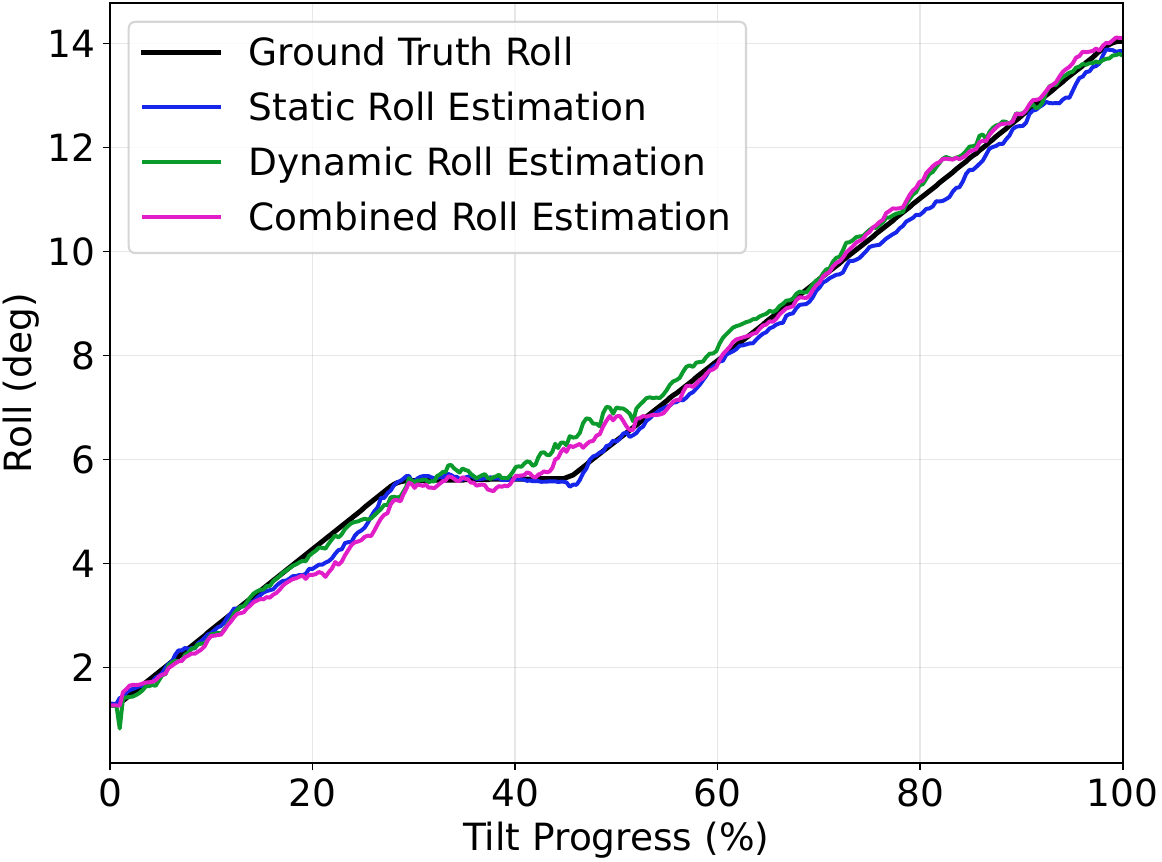}
        \label{fig:pause_roll_example}
    }

    \caption{Comparison between the estimated and ground-truth angle curves for an example trial under temporal interruption of motion. Left: pitch curve; right: roll curve.}
    \label{fig:pause_pitch_roll_example}
\end{figure}

When data points from trials containing stop phases were used during training, the static representation consistently performed slightly better than the dynamic one, achieving the best mean MAE of $0.211^\circ$ in the motion phase and $0.251^\circ$ in the stop phase, while the combined representation did not outperform the static representation. Notably, when training with pause-inclusive trials that included stop-phase samples, the improvement in motion-phase angle regression was limited for all three representation settings. However, for angle regression during the stop phase, the inclusion of stop-phase data led to a more pronounced improvement for all representations, particularly for the dynamic and combined representation.

Fig.~\ref{fig:pause_pitch_roll_example} shows an example trial comparing the estimated angle curves of the three representations with the ground-truth angle curves. It can be seen that, during the intermediate pause phase, the estimated curves from the static representation were more stable and closer to the ground-truth curves. By contrast, the other two representations, especially the dynamic representation, exhibited more noticeable fluctuations both during the pause phase and shortly after it, possibly due to the reduced event rate during the pause.

\begin{figure}[!t]
    \centering

    \subfloat[Comparison of static contour regions under speed variation]{%
        \includegraphics[width=0.4\columnwidth]{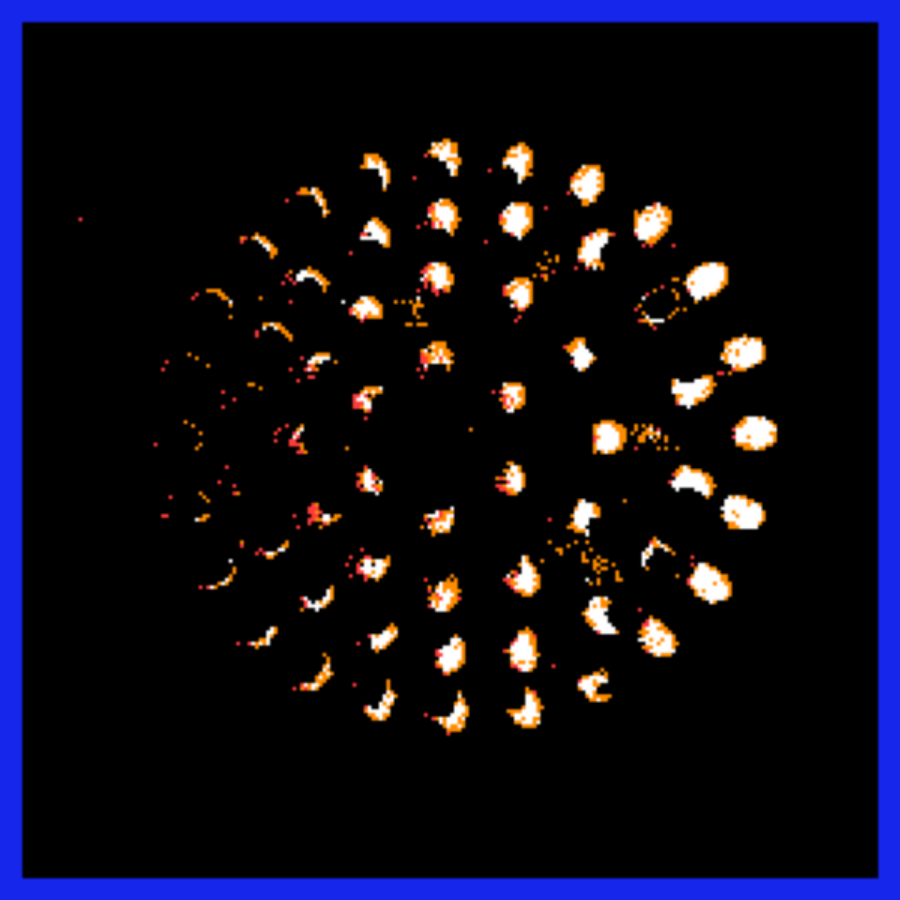}
        \label{fig:static_speed_overlay}
    }
    \hspace{0.02\columnwidth}
    \subfloat[Comparison of static contour regions under depth variation]{%
        \includegraphics[width=0.4\columnwidth]{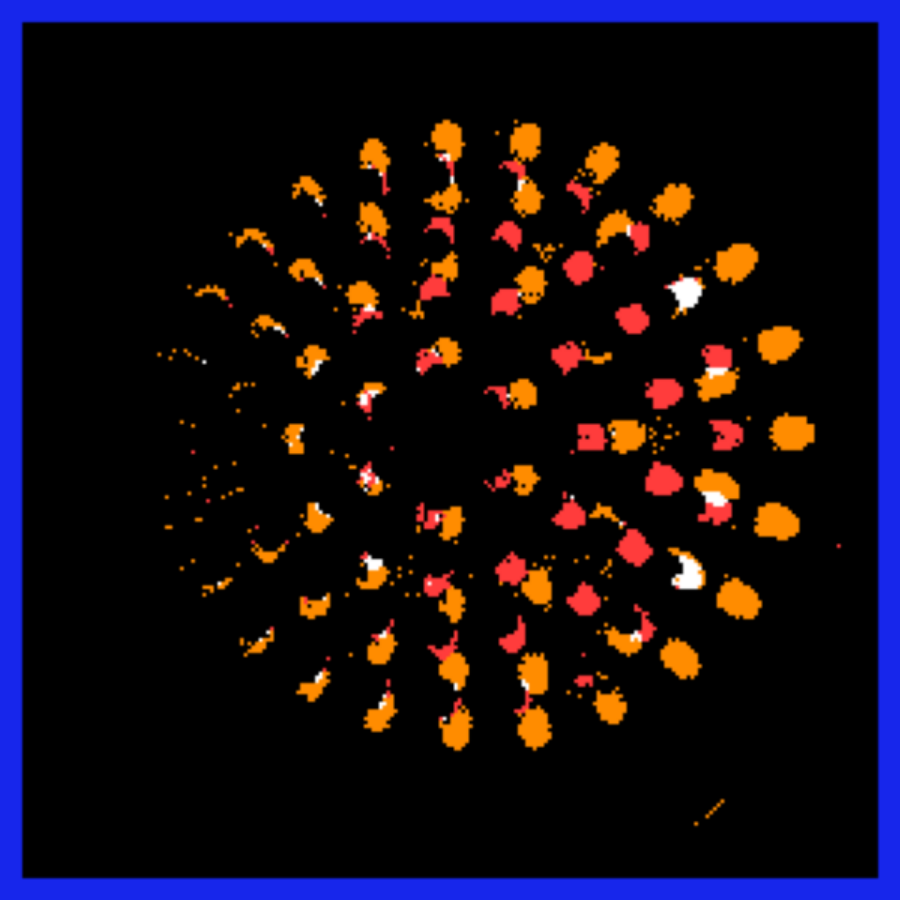}
        \label{fig:static_depth_overlay}
    }

    \vspace{0.6em}

    \subfloat[Comparison of dynamic contour regions under speed variation]{%
        \includegraphics[width=0.4\columnwidth]{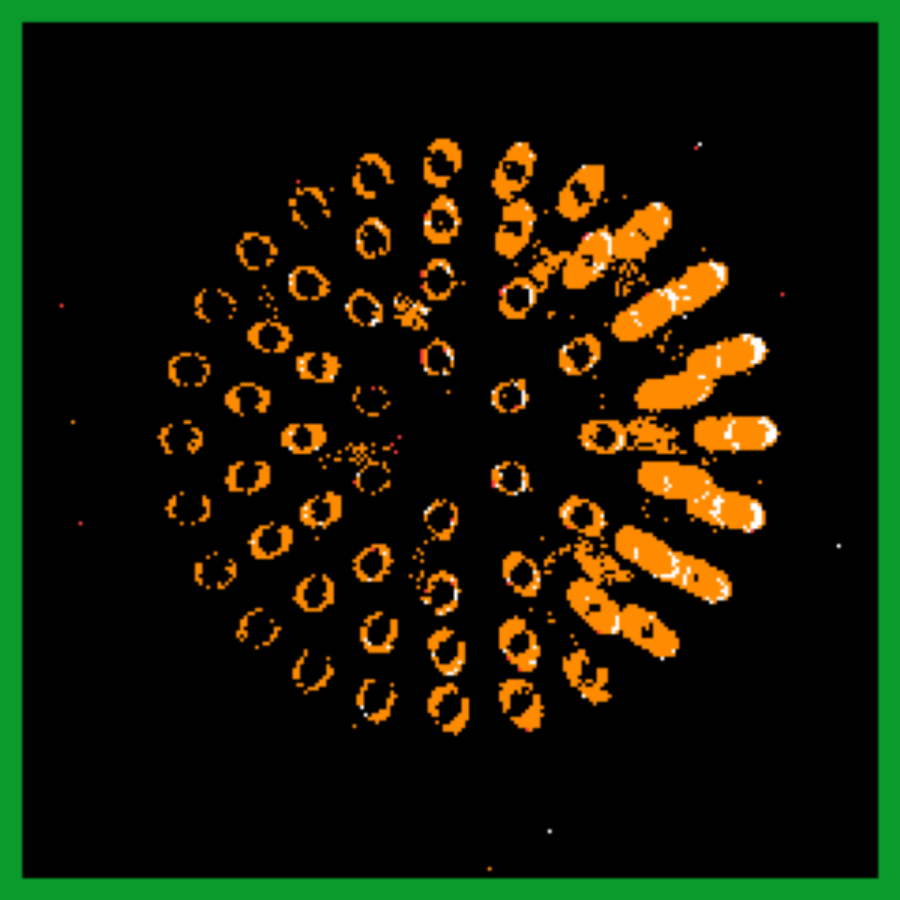}
        \label{fig:dynamic_speed_overlay}
    }
    \hspace{0.02\columnwidth}
    \subfloat[Comparison of dynamic contour regions under depth variation]{%
        \includegraphics[width=0.4\columnwidth]{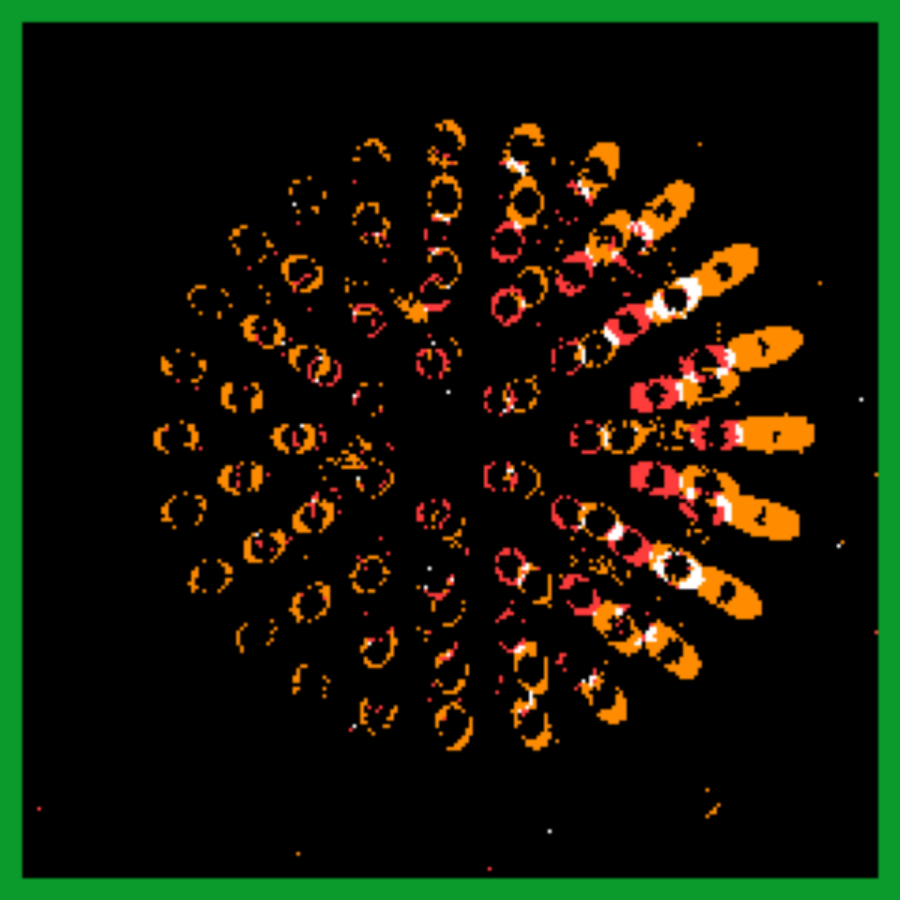}
        \label{fig:dynamic_depth_overlay}
    }

    \caption{Comparison of contour changes in the static and dynamic representations under speed and depth variations. For the speed comparison, the indentation depth is fixed at $2.0\,\text{mm}$, while for the depth comparison, the speed is fixed at $6\,^\circ/\text{s}$. Red contours indicate the cases with a speed of $2\,^\circ/\text{s}$ or a depth of $1.0\,\text{mm}$, orange contours indicate the cases with a speed of $10\,^\circ/\text{s}$ or a depth of $3.0\,\text{mm}$. Overlapping contours are shown in white.}
    \label{fig:static_dynamic_speed_depth}
\end{figure}

\subsection{Effect of speed and depth variations on angle regression}

Fig.~\ref{fig:static_dynamic_speed_depth} presents representative examples of the recovered binary spatial contours under different sensor rolling speeds and indentation depths for the two individual representations. The examples visually illustrate that the static representation remains relatively similar under speed variation, whereas the dynamic representation appears more sensitive to speed variation, showing less clear contour edges at $2\,^\circ/\text{s}$ and a more noticeable trailing effect at $10\,^\circ/\text{s}$. They also suggest that increasing the indentation depth enlarges the spacing and area of the recovered contours for both representations, with a more pronounced change in contour area for the dynamic representation.

\begin{figure}[!t]
    \centering

    \subfloat[]{%
        \includegraphics[width=0.48\linewidth]{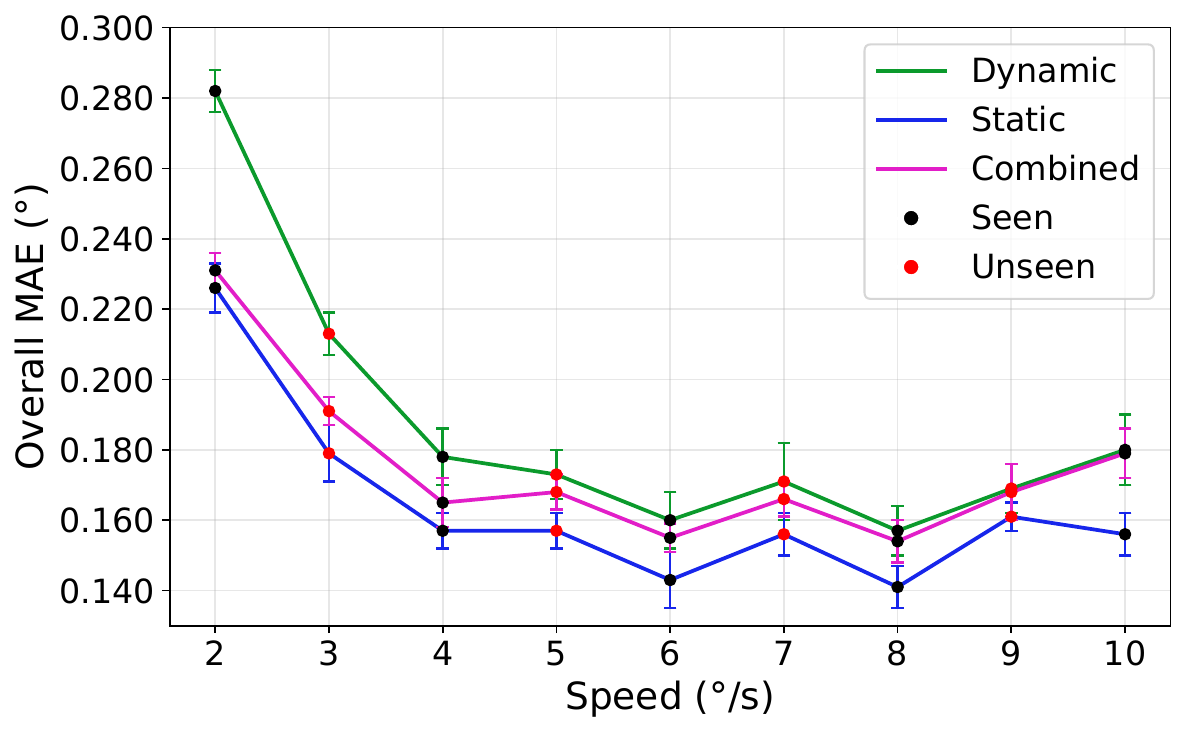}
        \label{fig:error_vs_speed}
    }
    \hfill
    \subfloat[]{%
        \includegraphics[width=0.48\linewidth]{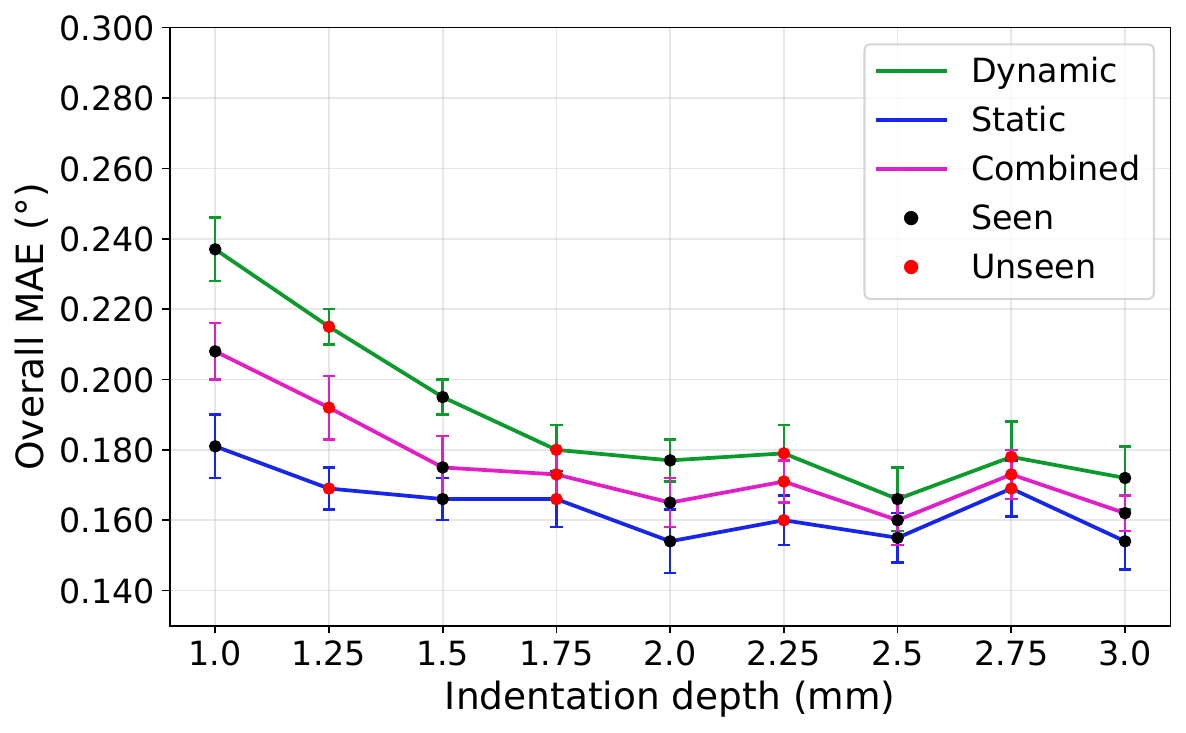}
        \label{fig:error_vs_depth}
    }

    \caption{Regression errors of different representations under speed and indentation depth variations during continuous motion, reported as mean and standard deviation over ten independent runs. Left: regression errors under speed variation; right: regression errors under depth variation.}
    \label{fig:speed_depth_prediction_results}
\end{figure}

For speed variation, Fig.~\ref{fig:speed_depth_prediction_results} shows that the static representation consistently achieved the lowest or tied-lowest mean regression error across all speed conditions, indicating the most stable performance. All three representations exhibited larger mean errors at the very low speed of $2\,^\circ/\text{s}$, with the increase being more pronounced for the dynamic representation, suggesting that dynamic cues are more strongly affected when motion-induced event activity is insufficient. Once the speed exceeded $4\,^\circ/\text{s}$, the mean error variations became smaller for all three representations. Moreover, the unseen interpolated speeds did not cause a systematic increase in mean error, as their performance generally remained close to that of the neighboring seen speeds, indicating interpolation generalization.

For indentation depth variation, the relative ranking of the three representations remained unchanged. The static representation showed smaller performance fluctuations across depths, whereas the dynamic representation was more sensitive to depth changes, especially at shallow indentation depths, where the weaker contact produced a lower event rate. As with speed variation, the unseen interpolated depths did not cause abrupt mean error increases, indicating interpolation generalization.

\subsection{Analysis of representation pipeline processing time}

\begin{figure}[!ht]
    \centering
    \begin{minipage}[b]{0.32\linewidth}
        \centering
        \includegraphics[width=\linewidth]{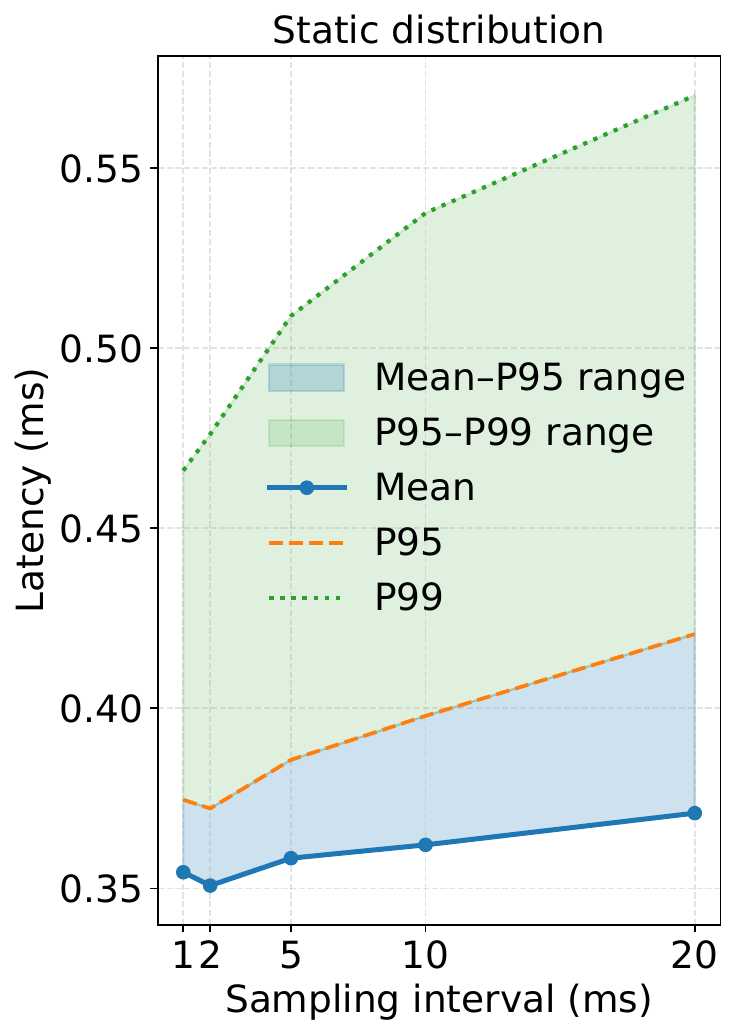}
    \end{minipage}
    \hfill
    \begin{minipage}[b]{0.32\linewidth}
        \centering
        \includegraphics[width=\linewidth]{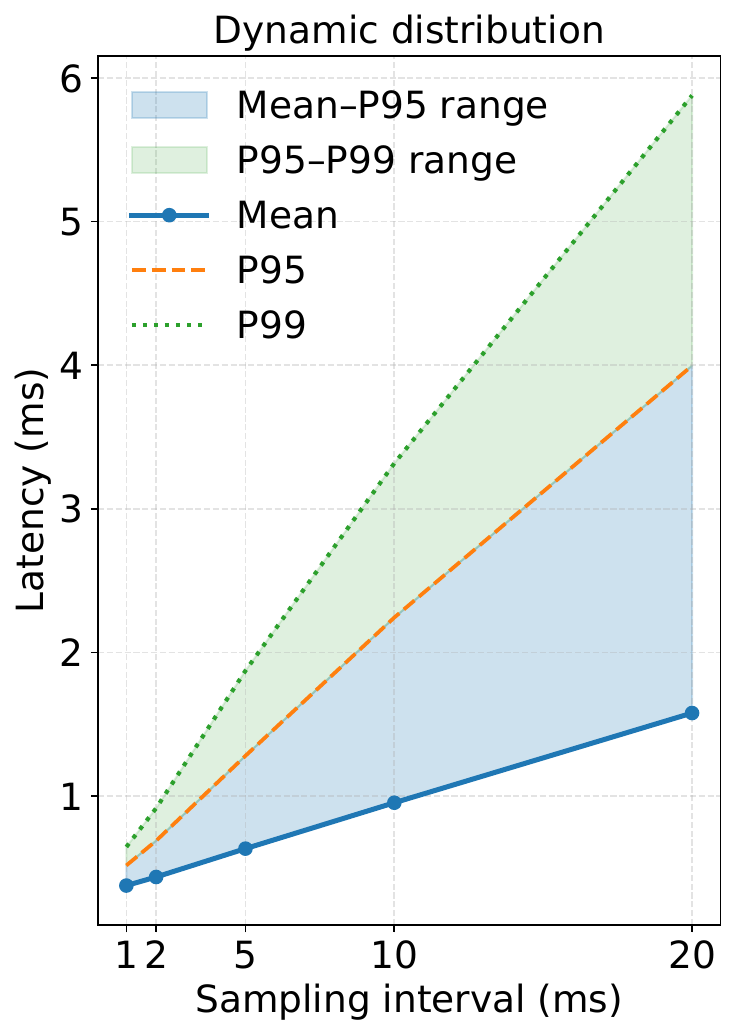}
    \end{minipage}
    \hfill
    \begin{minipage}[b]{0.32\linewidth}
        \centering
        \includegraphics[width=\linewidth]{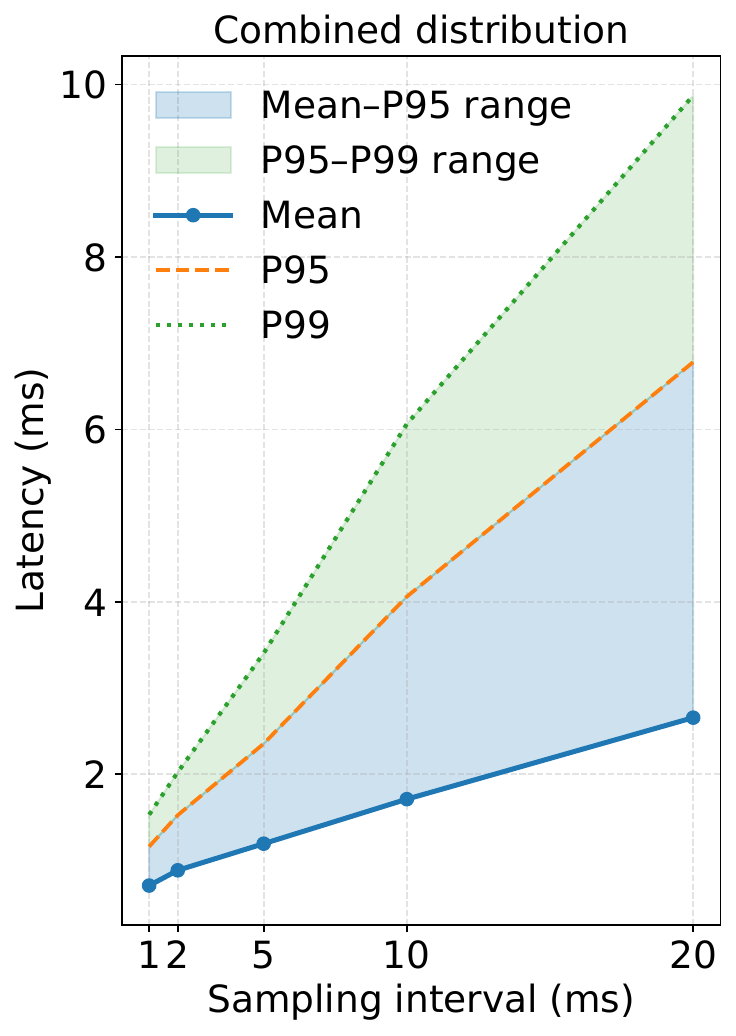}
    \end{minipage}

    \caption{Processing time distributions of the static, dynamic, and combined representation pipelines under different prediction intervals.}
    \label{fig:processing_time_distribution}
\end{figure}

When processed under different sampling intervals, all three representations showed longer overall pipeline times at larger intervals because more events were included in processing. Using an NVIDIA GeForce RTX 2080 Ti for network inference and an Intel Core Ultra 7 165H for all other processing, Fig.~\ref{fig:processing_time_distribution} shows the end-to-end processing time distributions from event-data input to the end of network inference. Excluding sensor-to-host transmission latency, the processing times of both the static and dynamic representations remained shorter than the corresponding sampling interval across all tested settings, whereas those of the combined representation exceeded the $1\,\text{ms}$ and $2\,\text{ms}$ interval for a small number of samples despite parallel branch generation. Nevertheless, the processing times of all three representations remained below $10\,\text{ms}$ across all tested sampling intervals.

The static representation showed lower latency partly because it used the dv-processing event accumulator, whose core operations are implemented with an efficient C++ backend, whereas the dynamic representation relied mainly on Python-based sliding-window updates. This also explains the higher latency of the combined representation and suggests that further optimization of the underlying implementation could reduce the latency of dynamic-related pipelines. The static representation required relatively more processing, as accumulation had to start from the first contact and continue thereafter. By contrast, the dynamic representation required initialization of the count frame over the initial window before estimation; however, because event activity was low before motion onset, this initialization introduced no significant additional delay.

\section{Discussion}

This study systematically compares static, dynamic, and combined event-derived spatial representations for tactile contact-angle regression during continuous contact posture variation. Overall, across both unidirectional continuous rolling and continuous rolling with randomly inserted pauses, the static representation achieved slightly better performance than the dynamic representation, likely because it recovers a more stable state-oriented spatial contour that is more beneficial for angle estimation than short-term dynamic changes alone. However, pixel-wise OR fusion of the two representations did not further improve the regression performance when the models were trained and evaluated within the same motion scenario. Across different rolling speeds and indentation depths, the performance of the static representation varied less than that of the dynamic representation. When the event rate decreased, the dynamic representation tended to be more susceptible to performance degradation.

Our static representation achieved a mean overall MAE of $0.160^\circ$ on the continuous-rolling test set. For trials containing an intermediate pause, it yielded a motion-phase mean MAE of $0.211^\circ$ and a stop-phase mean MAE of $0.251^\circ$. These results fall within the range reported in previous vision-based tactile angle estimation studies ($0.13^\circ$--$0.63^\circ$)~\cite{halwani2024novel,zaid2022elastomer,sajwani2023tactigraph}, and the pipeline processing times of the different representations indicate their potential for low-latency processing.

At the same time, this study found that, although the sensor was set to roll about the center of the tip sphere, the internal silicone exhibited elastic recoil when intermediate pauses were introduced. This may be related to stress redistribution around the pause position, which introduces distinct data patterns and reduces the adaptability of models trained only on continuous-motion data. Future work could explore whether incorporating motion history information could further improve angle estimation under such conditions. Moreover, this study was constrained to a highly controlled single-direction tilting setting for accurate label acquisition. The underlying mechanisms may become more complex when both tilt direction and magnitude vary simultaneously, especially for the dynamic representation, and therefore require further study using more flexible label acquisition methods.

More generally, the relative advantage of static and dynamic event-based pipelines is likely to be task-dependent rather than universal. Although the static pipeline was more effective for angle regression in this study, dynamic representations may still be worth exploring for transient-contact tasks such as slip detection, motivated by the role of rapidly adapting receptors in encoding minute skin motion for grip control~\cite{johnson2001roles}. The lack of improvement from the current fusion method also does not rule out more structured fusion strategies. Inspired by tactile population coding, where combining multiple receptor classes can improve information transmission~\cite{corniani2022population}, future work could preserve static and dynamic channels separately and fuse them at the feature or decoder level using learnable weighting or attention mechanisms.

Future work could also move beyond the current two-channel static--dynamic design toward more fine-grained event-based tactile pipelines. For example, a previous study has separated dynamic touch signals into RA-I-like and RA-II-like channels, with RA-II-like vibration features encoding harmonic structure useful for speed-invariant texture perception~\cite{pestell2022artificial}. Similarly, in human tactile perception, SA-I and SA-II receptors differ in receptive-field properties and support different aspects of static contact sensing~\cite{johnson2001roles}. Such channel specialization may improve information efficiency and task-specific tactile perception.

Overall, these findings provide useful guidance for designing event-based tactile representations for contact-angle estimation under continuous contact posture variation and highlight the potential of lightweight spatial-structure recovery pipelines for robust, low-latency contact-state perception in robotic manipulation.

\balance
\bibliographystyle{IEEEtran}
\bibliography{main}

\end{document}